\documentclass{article}

\usepackage[final]{nips_2016}

\usepackage[utf8]{inputenc} 
\usepackage[T1]{fontenc}    
\usepackage{hyperref}       
\usepackage{url}            
\usepackage{booktabs}       
\usepackage{amsfonts}       
\usepackage{nicefrac}       
\usepackage{microtype}      
\usepackage[pdftex]{graphicx}

\title{A wake-sleep algorithm for recurrent, spiking neural networks}

\author{
  Johannes C.~Thiele\thanks{to whom correspondence should be addressed} \\
  Physics Master Program\\
  ETH Zurich \\
  \texttt{johannes.thiele@alumni.ethz.ch} \\
  \And
  Peter U.~Diehl \\
  Institute of Neuroinformatics \\
  ETH Zurich and University of Zurich\\
  \And
  Matthew~Cook \\
  Institute of Neuroinformatics \\
  ETH Zurich and University of Zurich\\
}

\begin{document}

\maketitle

\begin{abstract}
We investigate a recently proposed model for cortical computation which performs relational inference. It consists of several interconnected, structurally equivalent populations of leaky integrate-and-fire (LIF) neurons, which are trained in a self-organized fashion with spike-timing dependent plasticity (STDP). Despite its robust learning dynamics, the model is susceptible to a problem typical for recurrent networks which use a correlation based (Hebbian) learning rule: if trained with high learning rates, the recurrent connections can cause strong feedback loops in the network dynamics, which lead to the emergence of attractor states. This causes a strong reduction in the number of representable patterns and a decay in the inference ability of the network. As a solution, we introduce a conceptually very simple ``wake-sleep'' algorithm: during the wake phase, training is executed normally, while during the sleep phase, the network ``dreams'' samples from its generative model, which are induced by random input. This process allows us to activate the attractor states in the network, which can then be unlearned effectively by an anti-Hebbian mechanism. The algorithm allows us to increase learning rates up to a factor of ten while avoiding clustering, which allows the network to learn several times faster. Also for low learning rates, where clustering is not an issue, it improves convergence speed and reduces the final inference error.

\end{abstract}

\section{Introduction}

The training of recurrent, spiking neural networks with spike-timing dependent plasticity (STDP) poses a big challenge for several reasons. Learning dynamics are often highly dependent on the parameters used for neurons and synapses and are difficult to control. Additionally, networks with a high level of recurrency and a correlation based STDP learning rule are susceptible to the emergence of feedback loops in the spiking dynamics when excitatory learning rates are high. This behavior can cause clusters in their weight matrices, which produce very strong attractors in the network dynamics. These problems are among the reasons why efforts in building functional spiking systems have mostly focused on feed-forward (\cite{Kheradpisheh:2016}, \cite{Masquelier:2007}) and shallow architectures (\cite{Bichler:2011}, \cite{Diehl:2016}), with highly abstract notions of recurrency (e.g. winner-takes-all inhibition), or the number of plastic recurrent connections was substantially reduced (see for instance \cite{Srinivasa:2012}, \cite{Wu:2008} and \cite{Davison:2006}). The issue of avoiding these feedback loops and the clusters they produce poses a serious limitation on the maximum learning rate which still allows convergence to a low inference error, and using low learning rates seems so far the only way to prevent recurrent spiking networks from developing these clustered structures. 

In this paper, we investigate the phenomenon of weight clustering on a network with a high number of recurrent connections at the core of its pattern recognition mechanism. As a possible solution of the aforementioned problem, we motivate and describe a conceptually simple ``wake-sleep'' algorithm (WSA), which uses the recurrent spiking dynamics of the network to identify attractors and the weights associated with them. By making use of the correlated activity of the neurons participating in the attractor state, these weights can be unlearned by a simple anti-Hebbian version of the normal excitatory learning mechanism. We demonstrate that the algorithm allows us to reliably prevent the appearance of clusters and enables us to increase the learning rate by up to a factor of 10. Additionally, the algorithm is able to remove clusters in the network if they are already present. Also for low learning rates, where clusters will usually not appear, the algorithm improves the learning speed and reduces the final inference error. 

Although our analysis is performed on only one specific network architecture, we believe that the general functional properties of the WSA are not constrained to this particular model, since it does not make use of any specific properties of the architecture besides its recurrent spiking dynamics. The problem of weight clustering directly stems from the Hebbian nature of the STDP learning scheme and therefore presents a very general problem when training recurrent spiking networks. This makes us believe that the WSA presented here could be applicable to every spiking network model where recurrent connections play a significant role in pattern representation. 

\section{Methods and problem statement}

This section describes the network model and the main components of the algorithm. All simulations were performed with the Brian spiking neural network simulator by \cite{Goodman:2008}. 

\subsection{Network and neuron model}

\begin{figure}[h]
  \centering
  \includegraphics[width=1\linewidth]{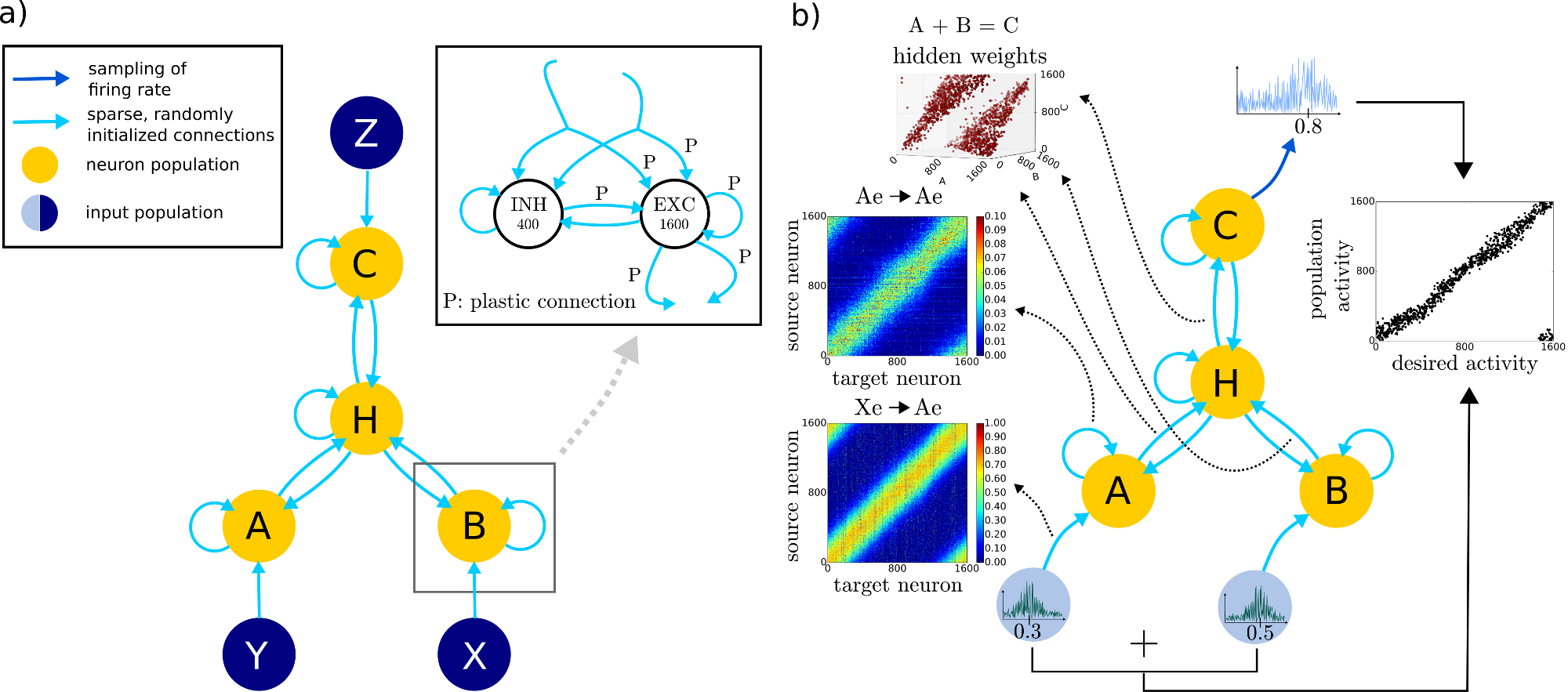}
  \caption{\textbf{\textsf{a)}} Visualization of the three-way architecture by \cite{Diehl:2016}. All neuron populations consist of 2000 leaky integrate-and-fire (LIF) neurons with 80\% excitatory and 20\% inhibitory neurons. Input populations are composed of 1600 Poisson-spike-generators which provide for $0.25\,\mathrm{s}$ the input as a Gaussian-shaped firing rate profile. All plastic connections are learned with STDP, using the triplet-rule by \cite{Pfister:2006} for exc. to exc. connections and inhibitory plasticity by \cite{Vogels:2011} for inh. to exc. connections. \textbf{\textsf{b)}} Weight matrix structure and inference of the network after successful learning of the relation $A+B=C$ (all numbers are constrained to the interval $[0,1)$ with periodic boundaries). The (resorted) input weights and recurrent weights (here shown for population $X$ and $A$) show a diagonal structure which reflects the distributed representation of bell-shaped input patterns over the neurons of population $A$. The hidden weights of population $H$ are visualized by assigning to each exc. neuron in $H$ a coordinate in a 3-dimensional space, which is spanned by the labels of the exc. neurons in the peripheral populations. Coordinates are assigned by finding that neuron in each peripheral population which has the strongest average connection strength to the corresponding hidden neuron. As we can see, the neurons in $H$ form a plane in this 3-dimensional space, which represents the solutions of the relation. During the testing (i.e. inference) phase, two peripheral populations receive a firing rate profile as during learning, while the firing rate profile of the exc. neurons of the third population is recorded and decoded with a population code to obtain the inferred value. All plasticities are disabled during this phase. \label{explain_weights}}
\end{figure}

We investigate our algorithm on a network performing relational inference, which was develop as a abstract model of cortical computation by \cite{Diehl:2016}. A short description can be found in figure \ref{explain_weights}. For details on the implementation, we refer the reader to the original publication. In this paper, we use the identical structure as in the original model and will only alter the learning rates of synapses between excitatory neurons. 
Note that the unique feature of this architecture, which distinguishes it from similar models (e.g.\ \cite{Srinivasa:2012}, \cite{Wu:2008}, \cite{Davison:2006}), is the high number of plastic recurrent connections. All connections are initialized with random sparse matrices, which imposes only little initial structure on the weights. Connections going to excitatory neurons (inh. to exc. and exc. to exc.) are plastic, while all other connections maintain their initial random values. Another interesting property of the model is its modularity, with every single population being able to learn patterns independently if isolated from the network. These independent modules are stacked together to form an architecture for relational inference. Figure \ref{explain_weights} shows the structure of the weight matrices after successful learning of the relation $A+B=C$.  

\subsection{Creation of weight clusters by recurrent feedback loops}

\begin{figure}[h]
  \centering
  \includegraphics[width=0.7\linewidth]{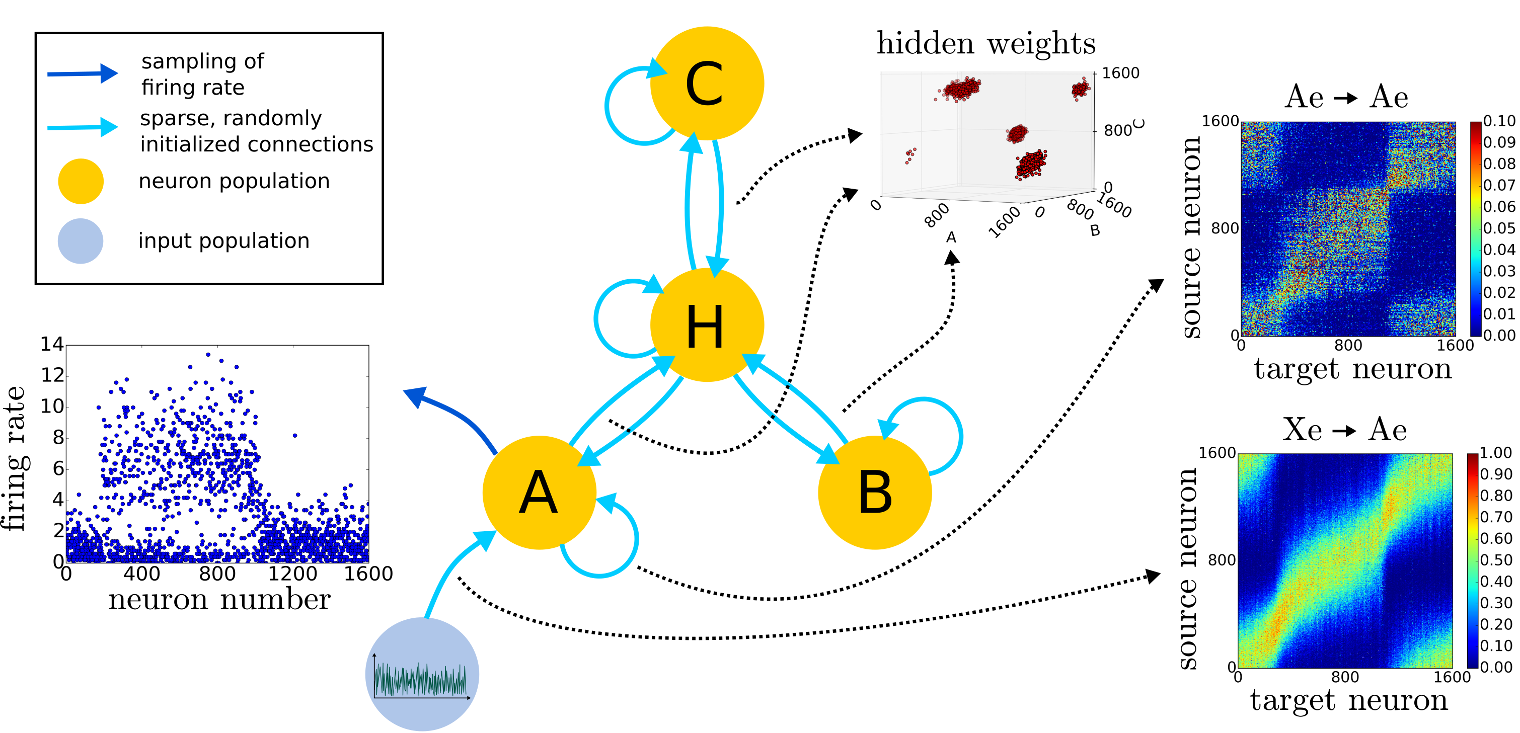}
  \caption{Activation of network attractors during the sleep phase. Random firing rates are provided via one of the input populations, which will activate one of the attractors present in the network dynamics. The firing rate pattern evoked by this random activity in population $A$ is shown in the lower left. The attractor state is clearly visible as a subgroup of neurons with elevated spiking activity, which consists precisely of those neurons which are interconnected by one of the block-like clusters in the recurrent weight matrix. By reversing the sign of the excitatory learning rates, the connections forming the clusters are weakened due to the high activity of the interconnected neurons (see also figure \ref{sleep_unlearning}). \label{generative_sampling_paper}}
\end{figure}

Figure \ref{high_nu_WSA} compares the result of the learning process with the standard learning rate for excitatory synapses and a tenfold increase in this learning rate. We can observe that this increase in learning rate leads to the appearance of clusters in the weight matrices of peripheral as well as hidden populations. These clusters cause correlated activity of a large subgroup of neurons, which leads to a strong attractor state in the dynamics of the network (see figure \ref{generative_sampling_paper} for a demonstration of the relation between recurrent weight clustering and firing dynamics). Figure \ref{high_nu_WSA} also demonstrates the effect of these clusters on the inference ability of the network. For a large number of possible input patterns, the network dynamics will converge to one of a few attractor states, which leads to a serious reduction of the number of possible answers the network can give to the inference task. 

The emergence of these clusters is caused by the Hebbian nature of the STDP learning mechanism. The strong recurrent dynamics of the network will lead to feedback loops in small subgroups of neurons which are more strongly interconnected. Since the STDP rule is based on the reinforcement of correlated activity, these feedback loops will strengthen connections between those neurons which are highly active in this feedback loop. If learning rates are low, this effect is balanced by continuously presenting new external input to the network, which makes sure that different neurons are activated for different patterns. However, if the learning rate is increased, the recurrent dynamics quickly reinforce the connections between those neurons which are already more active. This way the weights collapse into clusters and the recurrent dynamics are reinforced to a point where even very different input patterns will activate only a few distinct subgroups of neurons which are interconnected by such a weight cluster. The aim of our algorithm is to unlearn those weights which are responsible for this correlated activity. We show that this enables us to speed up learning of the network while avoiding clustering.

\subsection{The wake-sleep algorithm}

We now describe how a wake-sleep algorithm for spiking neuron populations can be implemented. The wake and sleep phase are defined as follows:
\begin{itemize}
\item \textit{Wake phase}: In this phase the network learns new patterns. Input is externally provided by firing-rate patterns of the input neurons, which encode the numbers representing solutions of the relation. All possible synaptic plasticities are enabled and on their standard values. The number of training examples shown during the wake phase is $r_\mathrm{sleep}$, which we call the sleep rate (since it describes after how many examples sleep begins).
\item \textit{Sleep phase}: During the sleep phase no external input patterns are presented to the network. Instead, the network will experience its own ``dreams'', which are the spiking dynamics induced in the network by a random input pattern provided through the input populations. The random input is produced by choosing for the firing rate of each input neuron a random number in the interval $[0,1]$ and then scaling all firing rates such that the total activity of the input population is equivalent to the total activity of the receiving peripheral population as observed during the wake phase. Our experiments with the algorithm have shown that it works best if during the sleep phase random input is only provided to one peripheral population at a time for a sleep time $t_\mathrm{sleep}$, with all other populations receiving no input (see figure \ref{generative_sampling_paper}). This is repeated for all other populations during the sleep phase (the order is chosen randomly). Additionally, we reverse the sign of all excitatory learning rates, which induces a form of anti-Hebbian plasticity. This will weaken the connections between all those neurons which are part of a feedback loop, reducing the influence of the attractor state on the network dynamics. We also apply every $0.25\,\mathrm{s}$ the weight normalization used as a homoeostatic mechanism in the original model (this is the same time for which a single example is presented during wakefulness). 
\end{itemize} 
Note that the mechanism does not require any changes in the network structure and the only parameter which has to be adapted compared to the wake phase is the sign of the excitatory learning rates. The dreams are produced by the generative model of the network, which is formed by its recurrent connections and the feedback loops they induce in the network dynamics. Providing random input to the network during the sleep phase will activate one of its attractors states, which is then unlearned by the anti-Hebbian mechanism. Figure \ref{sleep_unlearning} demonstrates the effect of the sleep phase on existing clusters in peripheral population $A$. We can observe that the recurrent connections of the weight clusters are indeed weakened during the sleep phase.    

The idea of an unlearning mechanism during sleep is not new: in their hypothetical theory about the ``function of dream sleep'', \cite{Crick:1983} proposed a very similar mechanism which unlearns attractor states (``dreams'') of the brain dynamics. These are activated by providing some form of random input during REM sleep. Although it lacks so far a convincing biological justification, the functional benefits of such an unlearning mechanism in recurrent networks have been demonstrated by \cite{Hopfield:1983}. Additionally, it can be related to the training algorithms of Boltzmann and Helmholtz machines (see \cite{Hinton:1983}, \cite{Rumelhart:1986} and \cite{Hinton:1995}), where the subtraction of the ``free-running'', generative contributions in the weight update rules can be interpreted as an ``unlearning'' of representations which are not related to the external input patterns (and therefore only ``dreams''). Our algorithm can thus be seen in the tradition of these well established wake-sleep algorithms.

\section{Results}

\subsection{Prevention of weight clustering and stabilization of network dynamics}

\begin{figure}[h]
  \centering
  \includegraphics[width=1\linewidth]{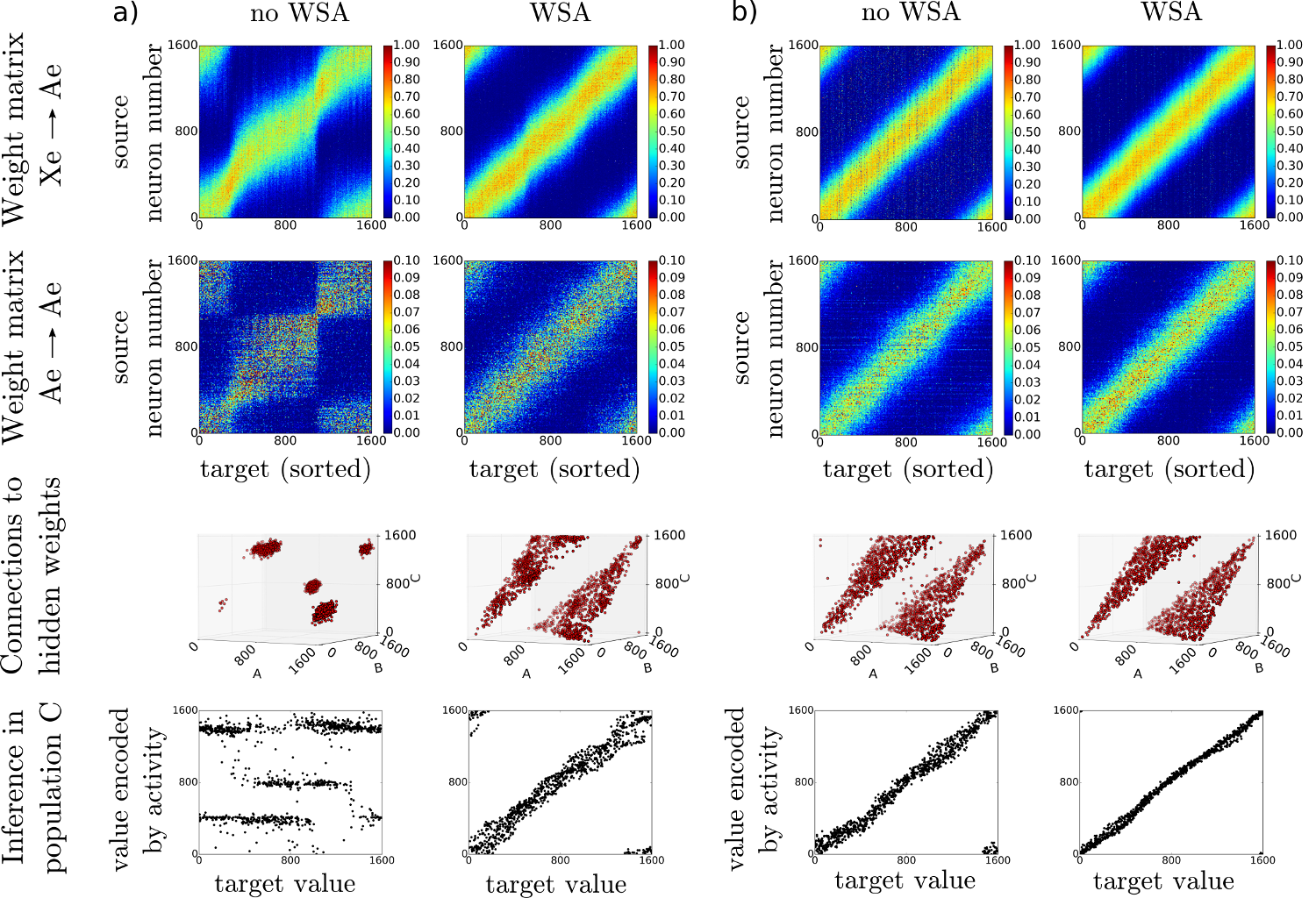}
  \caption{Effect of the WSA on training with high and low learning rates. The network is trained to learn the relation $A+B=C$. Please refer to figure \ref{explain_weights} for a more detailed description of the different plots. \textbf{\textsf{a)}} Structure of the weight matrices after learning with tenfold learning rate (15000 examples). If no WSA is used, the matrices of all weights collapse into clusters. When training with the WSA ($r_\mathrm{sleep}=100$ and $t_\mathrm{sleep}=5\,\mathrm{s}$) we can observe that it prevents clusters from appearing in the matrices while supporting the faster learning speed. The inference plot shows no discretization, which enables the network to infer correctly. \textbf{\textsf{b)}} Structure of the weight matrices after training of the standard configuration with and without WSA for 25000 examples. If the WSA is used for training, the distribution of neurons on the plane becomes more homogeneous, with less neurons being assigned to the wrong position in space. Also the response in the inference plot seems much sharper, which indicates a higher inference accuracy. This intuition is supported by the error evolution shown in figure \ref{compare_errors}. \label{high_nu_WSA}}
\end{figure}

Figure \ref{high_nu_WSA} shows the final weight matrices after convergence if the WSA is applied for training. The effect is clearly visible: in contrast to training without the WSA, the network shows basically no clustering in the weights and inference is performed correctly. Additionally, the algorithm improves inference for low learning rates, where clustering is usually not a problem. Figure \ref{sleep_unlearning} shows that if the WSA is applied on a network without clusters, it tends to focus the recurrent weights on the diagonal, which increases the selectivity of the network for a small number of input neurons. This could explain the very sharp response plot we observe when the WSA is applied to the network with low learning rates. Note that the correct inference is based solely on the population vector of the inferring population, i.e.\ it is not necessary for the network to reproduce exactly the bell-shaped input firing rate profiles to perform a correct inference. Focusing the recurrent weights on the diagonal means that the network internally assigns a higher importance to the parts of the pattern with the highest firing rate (which are those close to the encoded value), which can be seen as a form of high pass filtering that reduces the influence of neurons with firing rates too low to represent meaningful information (for this specific input pattern presentation time). 

\subsection{Improvement of inference performance}

\begin{figure}[h]
  \centering
  \includegraphics[width=1\linewidth]{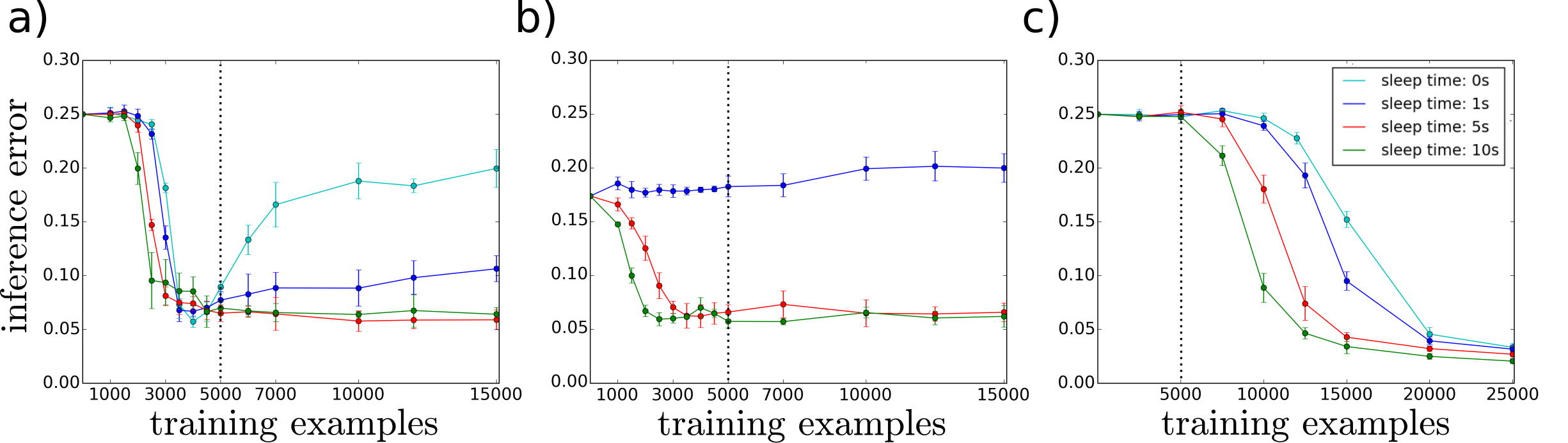}
  \caption{Comparison of inference errors for different excitatory learning rates and sleep times, using a sleep rate of $r_\mathrm{sleep}=50$.  \textbf{\textsf{a)}} Inference error for the $10$-fold learning rate: if the network is trained without the WSA ($t_\mathrm{sleep}=0$) the inference error drops initially and then starts increasing again. This sudden increase represents the point when the formation of clusters sets in. On the other hand, if the WSA is used for training, it prevents (for sufficiently high sleep times) reliably the formation of clusters, which improves the inference ability continuously with progressing training. \textbf{\textsf{b)}} Inference error for the $10$-fold learning rate when starting with clustered weight matrices. For sufficiently high sleep times, the WSA is able to remove weight clusters and enables convergence to a low inference error. Additionally, the inference error seems to decay even slightly faster than during training with random intial weights. \textbf{\textsf{c)}} Inference error for the standard configuration: we can see that in this case, the inference error is reduced continuously during training even if no WSA is applied. However, it can be seen clearly that the WSA significantly improves convergence speed and the final inference error also for low learning rates. The dashed line in all three plots visualizes the progress of training after 5000 examples. We can see that even compared to the standard model with WSA, a similar inference error is reached approximately twice as fast with high learning rates and WSA. \label{compare_errors}}
\end{figure}

Figure \ref{compare_errors} shows the evolution of the inference error as a function of the number of training examples for different learning rates and sleep times. For all analysed learning rates and sleep times, the WSA brings a significant improvement in inference accuracy and also reduces the number of examples which are necessary to achieve good inference performance. The effectiveness of the mechanism thus does not seem to depend on the exact amount of sleep, which indicates that the WSA is applicable without much fine-tuning of its parameters. 

\subsection{Unlearning of existing correlations}

\begin{figure}[h]
  \centering
  \includegraphics[width=1\linewidth]{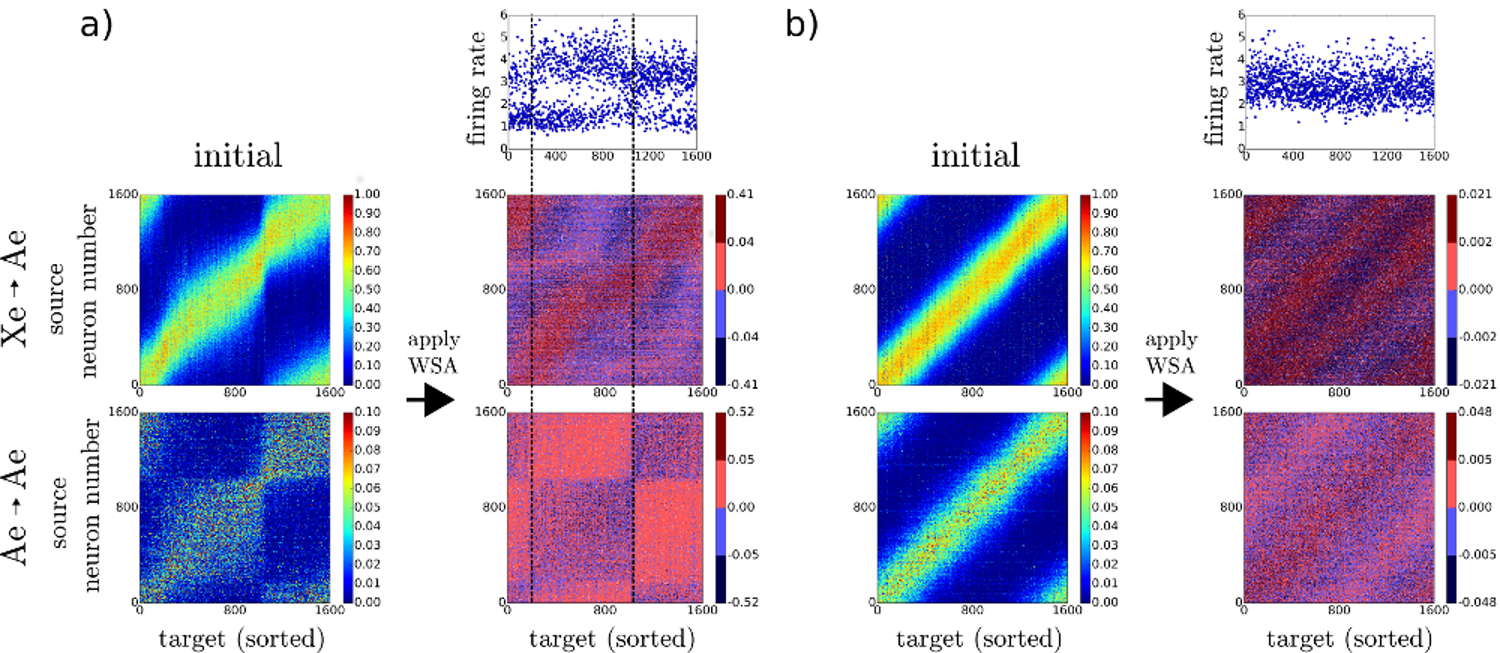}
  \caption{\textbf{\textsf{a)}} Visualization of the weight changes in population $A$ during the sleep phase in a network with weight clustering ($r_\mathrm{sleep}=100$ and $t_\mathrm{sleep}=5\,\mathrm{s}$). The images on the right of the arrow show the matrices of summed weight differences (sum over $20$ sleep phases). We can see clearly that the sleep phase induces a weight transfer from blocky areas to areas of weak synaptic weight for the recurrent weights. The input weights seem to be focused on the desired diagonal. The top shows the average firing rate of excitatory neurons over the $20$ sleep samples. The two attractor state are clearly visible as plateaus of elevated spiking activity of a subset of neurons. The dashed lines indicate the relationship between the firing rate of the neurons and the recurrent cluster they are part of. We can see that exactly those neurons which are recurrently interconnected by weight clusters seem to be proportionally more active during the sleep phase (compare with figure \ref{generative_sampling_paper} to see the effect for a single sleep phase). \textbf{\textsf{b)}} Visualization of the weight changes in population $A$ during the sleep phase in a network without clustered weights. The WSA leads to a weakening of the strong diagonal weights of the input weights, and reinforces the diagonal for the recurrent weights. It also seems that it makes the input weight diagonal slightly broader while concentrating the weights on the diagonal for the recurrent weights. Note that this is the opposite effect as observed for the clustered matrices, where the input weights are concentrated on the diagonal during the sleep phase. \label{sleep_unlearning}}
\end{figure}

\begin{figure}[h]
  \centering
  \includegraphics[width=1\linewidth]{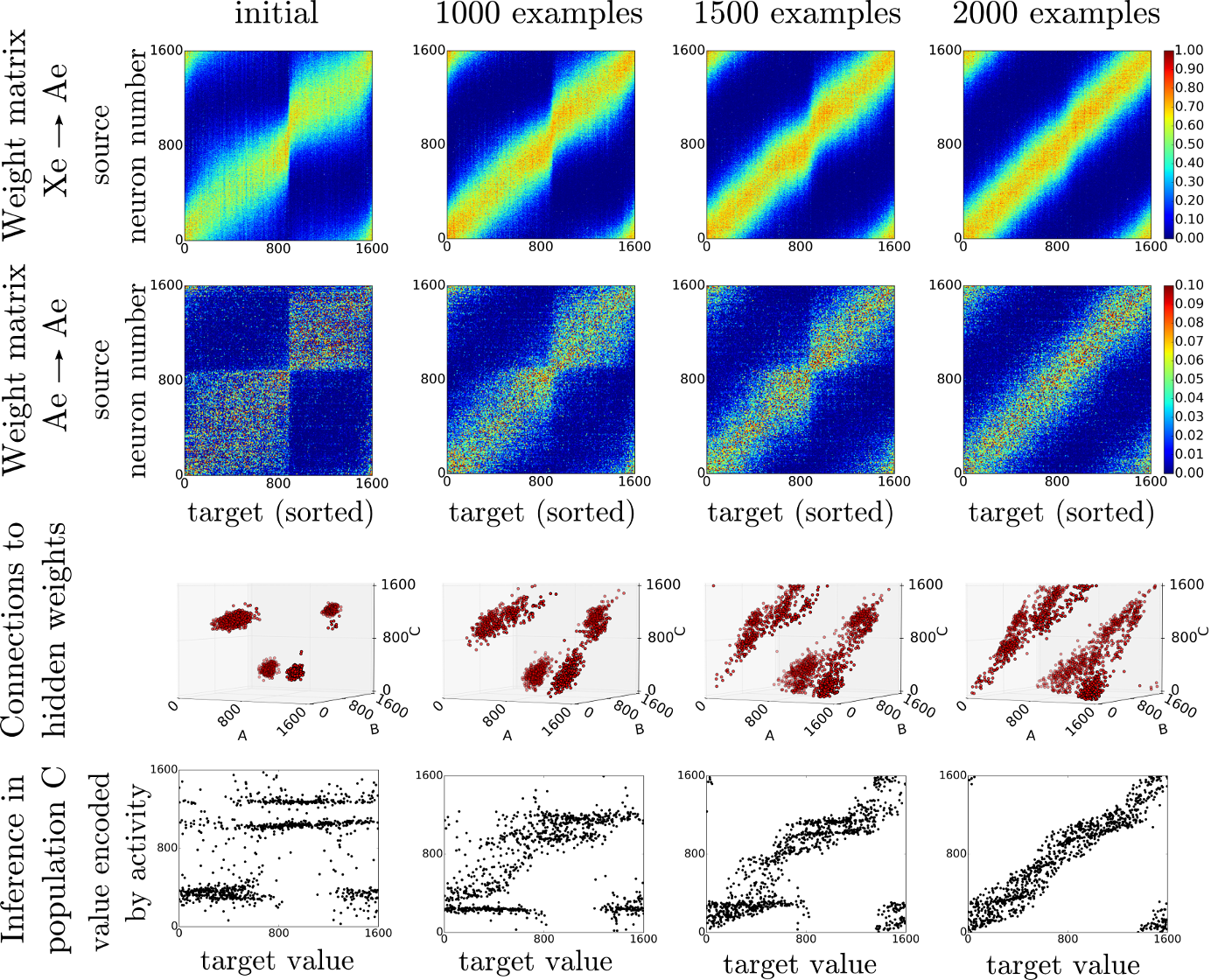}
  \caption{Time evolution of the weight matrices of a population which already shows strong clustering when continuing training with the WSA ($r_\mathrm{sleep}=50$ and $t_\mathrm{sleep}=5\,\mathrm{s}$). We can see that very quickly the clusters start to disappear in all weight matrices and the desired structure of the inference plot and weights is recovered. Note the quite small number of training examples after which the matrices are restored to their desired appearance again, which is even slightly faster than if we would have started with random initial weights (see also figure \ref{compare_errors}). \label{uncorrelate_high_nu}}
\end{figure}

In addition to preventing clusters from emerging in the network, the algorithm is able to remove clusters which are already present. Figure \ref{uncorrelate_high_nu} shows the result if weights with strong clusters are used as initial weights and the network is trained with high learning rate and WSA. It can be seen that all clusters start to disappear very quickly and the desired structure is recovered. In fact, convergence is even faster than with a random initial configuration (see figure \ref{compare_errors}). This demonstrates that the WSA possibly identifies exactly those weights for unlearning which are redundant, and leaves those weights mostly untouched which represent meaningful structures. This is supported by figure \ref{sleep_unlearning}, which shows the effect of the WSA on the weight matrices of peripheral population $A$. We can observe that for clustered weight matrices, the algorithm weakens the weights in the block-like clusters of the recurrent weights, while they are strengthened on the diagonal if no clusters are present. It therefore seems that the WSA leads the network to ``trust'' its internal representations (i.e.\ the recurrent dynamics) more than the external input if there are no clusters, while it tends to weaken the influence of the internal representations on the network dynamics (compared to the external input) if clusters exist.  

\section{Summary and Outlook}

It has been demonstrated how a high level of recurrency in spiking neural networks in combination with a correlation based STDP rule can lead to the appearance of clusters in the weight matrices for high learning rates. The correlated activity caused by these clusters can be understood as attractors in the network dynamics which have a detrimental effect on the inference ability of the network. 

We showed how random input patterns can be used to activate these attractors and identify the weight clusters which produce them. A simple anti-Hebbian mechanism, obtained by reversing the sign of the excitatory learning rates, can be used to unlearn precisely those weights which caused these attractor states. The wake-sleep algorithm developed from this concept was shown to be extremely successful in avoiding and removing clusters. It thus enables us to use much higher learning rates, which drastically reduces the number of training examples which are necessary to train the network to a low inference error. Also for low learning rates, it reduces the number of training examples necessary to achieve convergence and reduces the final inference error. We believe that the effectiveness of the mechanism stems from the fact that it makes use of the intrinsic recurrent dynamics of the network (which form its generative model) and therefore requires little fine-tuning if the network shows stable dynamics at least for low learning rates. This could imply that the WSA is also applicable to more general spiking network structures than the three-way network.

Although recurrent spiking network models are interesting from both a biological and a theoretical point of view, they have so far proven to be extremely difficult to train. Especially large-scale architectures (e.g.\ a scaled up version of the network used in this paper), with a possibly even higher number of recurrent connections, are vulnerable to cluster formation, which forces us to use low learning rates. The number of training examples necessary for convergence increases quickly with the number of neurons and the number of populations, in particular if populations lie ``deeper'' inside the network and receive input only indirectly via other populations. Additionally, scaling up the network further increases the level of recurrency and causes stronger feedback effects, which makes it more likely for the network to develop weight clusters and attractor states. Being able to access higher learning rates without causing this clustering is therefore extremely important to keep training time in a reasonable range for large-scale architectures. The wake-sleep algorithm presented in this paper was shown to be very successful in accomplishing this goal, which makes us believe that at least from a conceptual perspective, the algorithm could be a large step forward for our ability to train large-scale recurrent spiking network architectures. The investigation of its effectiveness on other recurrent architectures is therefore a promising direction for future research. Additionally, the algorithm was presented here in a minimal functional form, which still achieves the desired effect. Many details of the implementation leave room for further optimization, for instance a variation of learning and unlearning rates with progressing training or a mechanism which adapts the sleep time depending on the current level of weight clustering.

\section*{Acknowledgments}

P. Diehl and M. Cook were supported by SNF Grant 200021-143337. We are thankful for the useful discussions with the Cortical Computation Group at ETH Zurich/University of Zurich. We also want to thank ETH Zurich for granting its students access to the Euler computing cluster and CEA for supporting the presentation of this work at the NIPS 2016 workshop ``Computing with Spikes''.

\clearpage

\bibliography{references}

\end{document}